\documentclass[3p,times,fleqn]{elsarticle}
\makeatletter
\def\ps@pprintTitle{%
 \let\@oddhead\@empty
 \let\@evenhead\@empty
 \def\@oddfoot{}%
 \let\@evenfoot\@oddfoot}
\makeatother

\usepackage{amsmath}
\usepackage{amssymb}
\usepackage{psfrag}
\usepackage{color}
\usepackage{stfloats}
\usepackage{fixltx2e}
\usepackage{mathrsfs}
\usepackage{tabularx}
\usepackage{booktabs}
\usepackage{colortbl}
\usepackage{rotating}
\usepackage{boldline}
\usepackage{bm}
\usepackage{changes}
\usepackage{url}
\DeclareMathOperator*{\argmin}{arg\,min}

\usepackage{lineno}

\begin{document}
%%%%%%%%%%%%%%%%%%%%%%%%%%%%%%%%%%%%%%%%%%%%%%%%%%%%%%%%%%%%%%%%%%%%%%%%
\renewcommand{\topfraction}{0.98}	% max fraction of floats at top
\renewcommand{\bottomfraction}{0.98}% max fraction of floats at bottom
\setcounter{topnumber}{3}
\setcounter{bottomnumber}{3}
\setcounter{totalnumber}{4}         % 2 may work better
\setcounter{dbltopnumber}{4}        % for 2-column pages
\renewcommand{\dbltopfraction}{0.98}	% fit big float above 2-col. text
\renewcommand{\textfraction}{0.05}	% allow minimal text w. figs
\renewcommand{\floatpagefraction}{0.5}	% require fuller float pages
\renewcommand{\dblfloatpagefraction}{0.5}	% require fuller float pages
%%%%%%%%%%%%%%%%%%%%%%%%%%%%%%%%%%%%%%%%%%%%%%%%%%%%%%%%%%%%%%%%%%%%%%%%
\newcommand{\beq}{\begin{equation}}
\newcommand{\eeq}{\end{equation}}
\newcommand{\divg}{\mbox{\rm{div}}\,}
\newcommand{\Divg}{\mbox{\rm{Div}}\,}
\newcommand{\D}  {\displaystyle}
\newcommand{\DS} {\displaystyle}
\newcommand{\RM}[1]{\textit{\MakeUppercase{\romannumeral #1{}}}}
\newtheorem{remark}{\bf{{Remark}}}
\def\sca   #1{\mbox{\rm{#1}}{}}
\def\mat   #1{\mbox{\bf #1}{}}
\def\vec   #1{\mbox{\boldmath $#1$}{}}
\def\scas  #1{\mbox{{\scriptsize{${\rm{#1}}$}}}{}}
\def\scaf  #1{\mbox{{\tiny{${\rm{#1}}$}}}{}}
\def\vecs  #1{\mbox{\boldmath{\scriptsize{$#1$}}}{}}
\def\tens  #1{\mbox{\boldmath{\scriptsize{$#1$}}}{}}
\def\tenf  #1{\mbox{{\sffamily{\bfseries {#1}}}}}
\def\ten   #1{\mbox{\boldmath $#1$}{}}
\def\Ass  {\overset{\hspace*{0.4cm} n_{\scas{el}}}
          {\underset{\scaf{c},\scaf{d}=1}{\msf{A}}}}
\def\ltr   #1{\mbox{\sf{#1}}}
\def\bltr  #1{\mbox{\sffamily{\bfseries{{#1}}}}}
\sloppy
%%%%%%%%%%%%%%%%%%%%%%%%%%%%%%%%%%%%%%%%%%%%%%%%%%%%%%%%%%%%%%%%%%%%%%%%
\begin{frontmatter}
%%%%%%%%%%%%%%%%%%%%%%%%%%%%%%%%%%%%%%%%%%%%%%%%%%%%%%%%%%%%%%%%%%%%%%%%
\title{\Large Multi-fidelity classification using Gaussian processes: accelerating the prediction of large-scale computational models}

%\title{\Large Multi-fidelity classification using Gaussian processes: predicting the binary output of computer models using inexpensive approximations}

\author[01,02,03]{Francisco~Sahli Costabal$^*$}
\ead{fsahli1@uc.cl, corresponding author}
\author[04]{Paris Perdikaris}
\author[05]{Ellen Kuhl}
\author[01,02,03]{Daniel E. Hurtado}

\address[01]{Department of Structural and Geotechnical Engineering, School of Engineering, Pontificia Universidad Cat\'olica de Chile, Santiago, Chile}

\address[02]{Institute for Biological and Medical Engineering, Schools of Engineering, Medicine and Biological Sciences, Pontificia Universidad Cat\'olica de Chile, Santiago, Chile}

\address[03]{Millennium Nucleus for Cardiovascular Magnetic Resonance}

\address[04]{Department of Mechanical Engineering and Applied Mechanics, University of Pennsylvania, Philadelphia, Pennsylvania, USA}

\address[05]{Departments of Mechanical Engineering \& Bioengineering, Stanford University, California, USA}

%
%

%

%

%
%%%%%%%%%%%%%%%%%%%%%%%%%%%%%%%%%%%%%%%%%%%%%%%%%%%%%%%%%%%%%%%%%%%%%%%%
% an effective abstract is brief and normally less than 200 words.
% abstracts must not exceed 250 words - here: 393 words, cut in half!
%%%%%%%%%%%%%%%%%%%%%%%%%%%%%%%%%%%%%%%%%%%%%%%%%%%%%%%%%%%%%%%%%%%%%%%%
\begin{abstract} % 
Machine learning techniques typically rely on large datasets to create accurate classifiers. However, there are situations when data is scarce and expensive to acquire. This is the case of studies that rely on state-of-the-art computational models which typically take days to run, thus hindering the potential of machine learning tools. In this work, we present a novel classifier that takes advantage of lower fidelity models and inexpensive approximations to predict the binary output of expensive computer simulations. We postulate an autoregressive model between the different levels of fidelity with Gaussian process priors. We adopt a fully Bayesian treatment for the hyper-parameters and use Markov Chain Mont Carlo samplers. We take advantage of the probabilistic nature of the classifier to implement active learning strategies. We also introduce a sparse approximation to enhance the ability of themulti-fidelity classifier to handle large datasets. We test these multi-fidelity classifiers against their single-fidelity counterpart with synthetic data, showing a median computational cost reduction of 23\% for a target accuracy of 90\%. In an application to cardiac electrophysiology, the multi-fidelity classifier achieves an F1 score, the harmonic mean of precision and recall, of 99.6\% compared to 74.1\% of a single-fidelity classifier when both are trained with 50 samples. In general, our results show that the multi-fidelity classifiers outperform their single-fidelity counterpart in terms of accuracy in all cases. We envision that this new tool will enable researchers to study classification problems that would otherwise be prohibitively expensive. Source code is available at \url{https://github.com/fsahli/MFclass}.
\end{abstract}
%%%%%%%%%%%%%%%%%%%%%%%%%%%%%%%%%%%%%%%%%%%%%%%%%%%%%%%%%%%%%%%%%%%%%%%%
%%%%%%%%%%%%%%%%%%%%%%%%%%%%%%%%%%%%%%%%%%%%%%%%%%%%%%%%%%%%%%%%%%%%%%%%
\end{frontmatter}
%%%%%%%%%%%%%%%%%%%%%%%%%%%%%%%%%%%%%%%%%%%%%%%%%%%%%%%%%%%%%%%%%%%%%%%%%
%%%%%%%%%%%%%%%%%%%%%%%%%%%%%%%%%%%%%%%%%%%%%%%%%%%%%%%%%%%%%%%%%%

%\linenumbers

%figure 1: sine example
%figure 2: sine accuracy
%figure 3: spiral wave motivation
%figure 4: spiral wave results
%figure 5: spiral wave accuracy
%figure 6: multi-dimensional spiral wave
\section{Motivation}\label{intro}
%%%%%%%%%%%%%%%%%%%%%%%%%%%%%%%%%%%%%%%%%%%%%%%%%%%%%%%%%%%%%%%%%%%%%%%%%
Machine learning is a paradigm-shift field that is transforming many areas of engineering and sciences. Some of the most impressive applications of machine learning rely on the availability of large amounts of data that are used to create accurate classifiers that, for example, could discriminate between the absence or presence of a disease \cite{hannun2019cardiologist}. However, data required to train a classifier may not be available in a sufficient amount, or it can be and expensive and lengthy to acquire, thus hindering the potential of machine learning in many areas of application. This is the case in studies that largely rely on state-of-the-art computational models, which often take several days to run and require a significant amount of resources before one data point is obtained. Despite these restrictions, computational modeling represents an attractive alternative to physical experiments, which can be even more expensive, time consuming, and may pose important restrictions due to ethical considerations. Several areas of the sciences and engineering are already benefiting from computational modeling, including the medical device industry \cite{morris2016computational, lee2018propagation}, car and aircraft manufacturing \cite{forrester2008engineering}, and public policy design \cite{santner2003design}, to name a few. Despite their massive adoption, researchers can often only afford to perform a few runs of complex models that can have hundreds of parameters \cite{SahliCostabal2018}. 
%state-of-the-art computational models still take days to run and often have hundreds of parameters that need fitting \cite{SahliCostabal2018}. 
This introduces a bottleneck for understanding the parameter influence and the uncertainties associated with the model. Nonetheless, there are usually lower fidelity approximations of the model that are less expensive to compute and can provide valuable information. For example, the finite element method is widely used to solve problems in solid and fluid mechanics. Here, a natural low fidelity model would be a model with a coarser mesh. Although this approximation may not accurately represent the target quantities of interest, it will likely capture how parameters changes are reflected in the output. Another alternative for low-fidelity models is to reduce the dimensionality of the problem, replacing a fully three-dimensional problem with a two- or one-dimensional representation \cite{sahli2019machine}. Finally, experimental data can be considered as an additional information source with variable fidelity, either high or low, depending on the application. In recent years, there has been an increased attention in the machine learning community to develop predictive methods that enable the effective fusion of variable fidelity information sources. Some techniques are specially tailored to estimate the uncertainty in the model predictions  \cite{schiavazzi2017generalized,Hurtado2017,biehler2015towards,PWG17MultiSurvey,QPOVW17MFGSA} and parameter estimation \cite{koutsourelakis2009multi}. To predict the model output, Kennedy and O'Hagan \cite{kennedy2000predicting} first proposed to use Gaussian process priors to perform multi-fidelity regression. This approach has been widely used \cite{parussini2017multi} and has been extended for efficiency \cite{le2014recursive} and to accommodate large data-sets \cite{perdikaris2016multifidelity}. However, all these techniques have been solely developed in the regression setting, and none of them has yet been extended to classification problems.

There are many problems that would benefit from multi-fidelity tools for classification. Phase diagrams and most types of bifurcations in physical systems can be treated as a classification problem. Here, we will determine which state will be reached under a given set of parameters, initial or boundary conditions. A good example of this type of behavior is present in cardiac tissue dynamics. Due to the nature of excitable cells, under certain conditions, cardiac tissue can achieve arrhythmic states that are self-sustained, such as spiral waves \cite{gray1998spatial}. Another example comes from the mechanical analysis of bi-layered systems, such as the white and grey matter in the brain. Here, it is hypothesized that, during development, different rates of growth in the two layers causes a buckling instability that could lead to the formation of either creases or ridges \cite{budday2014mechanical,holland2017instabilities}.

In the Gaussian process setting, there are some fundamental differences between the classification and regression tasks that make the former harder to implement. Most notably, in the regression setting, a Gaussian likelihood is typically assumed for the data. Combined with a Gaussian process prior over the latent generating function, this allows for deriving an analytic expression for both the marginal likelihood and the predictive posterior of the regression model \cite{rasmussen2006gaussian}. In the classification setting, the Gaussian likelihood assumption is not appropriate for modeling discrete class labels, and it is not possible to formulate an exact inference scheme for training the model parameters. Therefore, we must resort to approximate inference techniques, such as the widely used Laplace approximation and the expectation propagation algorithm  \cite{rasmussen2006gaussian}. Markov-chain Monte Carlo (MCMC) methods provide a powerful alternative to approximate posterior inference, but standard sampling schemes such as Gibbs and Metropolis-Hastings suffer from slow convergence rates due to strong correlations in the Gaussian process posterior \cite{robert2013monte}. However, recent advances in Hamiltonian Monte Carlo have yielded improved sampling schemes that can perform well in this setting \cite{neal1999regression}. In this work, we leverage recent developments in the machine learning community, such as automatic differentiation and parallelization, to implement a novel efficient multi-fidelity classifier using Gaussian process priors. We will also present a sparse version to enhance the computational expediency of our method for large data-sets. We will demonstrate its superior performance against single-fidelity classifiers with synthetic data, as well as its effectiveness on a realistic large-scale application to cardiac electrophysiology. Since Gaussian process classifiers have an associated variance with their predictions, they naturally enable the design of effective data acquisition policies via active learning \cite{kapoor2007active,Gramacy2017}. In this way, it is possible to optimize the sampling strategy of expensive computer models to maximize accuracy under a constrained computational budget. We will show how this methodology is implemented in our multi-fidelity classifier and how it compares against standard single-fidelity classification models.

This paper is structured as follows: in Section~\ref{sec:methods} we introduce the multi-fidelity Gaussian process classifier, the sparse approximations and the active learning criteria, in Section~\ref{sec:synthetic} we test the classifier with a synthetic example to evaluate its performance, in Section~\ref{sec:cardic_ep} we use the classifier to identify a temporal region where an arrhythmia can be induced in cardiac tissue. We finalize this work discussing these results in Section~\ref{sec:discussion}.

\section{Multi-fidelity classification with Gaussian process}
\label{sec:methods}
We start this section by revisiting classification with Gaussian processes to form the base of the proposed multi-fidelity classifier. We later introduce the sparse approximations, the statistical inference technique we employ to train the multi-fidelity model, as well as the active learning approach for adaptive data acquisition.

\subsection{Single-fidelity classifier}
\label{sec:sf}

We start by assuming that we have a dataset comprised of input/output pairs $\mathcal{D} = \left\{ (\vec{x}_i, y_i)^N_{i=1} \right\} = \left\{\ten{X}, \vec{y}\right\}$. The inputs $\ten{X} \in \mathbb{R}^{N\times D}$ contain $D$ features for $N$ training examples and the outputs $\vec{y}$ contain the $N$ discrete class labels for the corresponding inputs. Here, we restrict the scope of this work to binary classification, thus the labels can only take on two values $y_i = \{0,1\}$. We note that it is possible to extend this framework to the multi-class classification setting. The classical formulation of Gaussian process classification defines an intermediate variable which is computed from a latent function $f(\vec{x})$ \cite{rasmussen2006gaussian}. Throughout this paper, we will assume standardized data-sets and work with zero-mean Gaussian process priors of the form $f \sim \mathcal{GP}\,(\vec{0},k(\vec{x},\vec{x}'; \theta))$. Here, $k(\cdot,\cdot;\theta)$ is a covariance kernel function, which depends on a set of parameters $\theta$. We adopt a fully Bayesian treatment and prescribe prior distributions over these parameters, which we will specify later \cite{neal1999regression}. To obtain class probability predictions we pass the Gaussian process output $f$ through a nonlinear warping function $\sigma(f)$, such that the output is constrained to [0,1], rendering meaningful class probabilities. We define the conditional class probability as $\pi(\vec{x}) = p(y =  1 | \vec{x}) = \sigma(f(\vec{x}))$. A common choice for $\sigma(f)$ is the logistic sigmoid function $\sigma(f) = (1 + \exp{(-f)})^{-1}$, which we will use throughout this work. We assume that the class labels are independent according to a Bernoulli likelihood with probability $\sigma(y)$ \cite{Nickisch2008}. Further, we choose an automatic-relevance determination (ARD) squared exponential kernel \cite{rasmussen2006gaussian}, $k(x, x'; \theta) = \eta \, \mathrm{exp}\left[-\sum_{m=1}^{M}(x_m - x_m')^2/(2 \ell_m^2) \right]$, parametrized by $\theta:=\{\eta, \ell_1,\dots,\ell_M\}$. We assume the following distributions for the parameters $\eta$ and $\ell_m$:
\begin{eqnarray}
  \eta & \sim & {\rm HalfNormal}(\sigma = 5)\\
  \ell_m & \sim & {\rm Gamma}(\alpha = 2, \beta = 2), \ \ m=1,\dots,M.
\end{eqnarray}
The posterior distribution over the model parameters $\theta$ cannot be described analytically, and thus we resort to approximate-inference techniques to calibrate this model on the available data. To this end, we use the NO-U-Turn sampler (NUTS) \cite{hoffman2014no}, which is a type of Hamiltonian Monte Carlo algorithm, as implemented in PyMC3 \cite{salvatier2016probabilistic}. We use two chains, and set the target accept probability between 0.95 to 0.99 depending on the convergence. The first 1000 samples are used to adjust the step size of the sampler, and are later discarded. We use the subsequent 1000 samples to estimate the parameters $\theta$. Once we have completed the inference, we can make predictions $\vec{y}^*$ at new location $\vec{x}^*$ in three steps: First we compute the predictive random variable $f^*(\vec{x}^*) \sim \mathcal{N}(\mu(\vec{x}^*),  \Sigma(\vec{x}^{\ast}))$, which by construction follows a multivariate normal distribution, with mean $\vec{\mu}$ and covariance $\Sigma$ obtained by conditioning on the available data \cite{rasmussen2006gaussian}:
\begin{eqnarray}
  \mu(\vec{x}^{\ast})  & =  & k(\vec{x}^{\ast}, \vec{X}) \vec{K} ^{-1}\vec{f} 
\label{eq:posterior_mean} \\
  \Sigma(\vec{x}^{\ast}) & = & k(\vec{x}^{\ast}, \vec{x}^{\ast}) - k(\vec{x}^{\ast}, \vec{X}) \vec{K}^{-1} k(\vec{X},\vec{x}^{\ast}) \, ,
\label{eq:posterior_variance}
\end{eqnarray}
where the covariance matrix $\vec{K}\in\mathbb{R}^{N\times N}$ results from evaluating the kernel function $k(\cdot,\cdot;\theta)$ at the locations of the input training data $\vec{X}$ and $\vec{f} = f(\ten{X})$. Then we sample $f^*$ drawing parameters from the estimated posterior distributions. This will result in a number of realizations of $f^*$ that we finally pass through the logistic sigmoid function $\sigma$ to obtain a distribution of class probabilities $\vec{y}^*$. Finally, we take the mean of this distribution to make a single prediction. 

\subsection{Multi-fidelity Gaussian process classification}

We assume that we have $s$ information sources, each with a different level of evaluation cost and fidelity, producing a class label $t_i$, with $i = \{1,s\}$. Here, level $s$ represents the highest, most accurate and expensive information source and 1 represents the most simplified and cheapest model available. We propose to model the cross-correlation structure between different levels using an auto-regressive model for the latent functions $f_i$ \cite{kennedy2000predicting}:
\begin{eqnarray}
f_i(\vec{x}) &=&  \rho_{i-1}f_{i-1}(\vec{x}) + \delta_i(\vec{x}), \hspace{1mm}i = 2,\dots,s, \label{eq:prior_f_i}\\
f_1(\vec{x}) &=& f_1(\vec{x}) \label{eq:prior_f_1}
\end{eqnarray}
Here, $\rho_i$ is a scalar parameter that needs to be inferred and $\delta_i$ is a function that aims to capture the bias in the predictions of the lower fidelity models. The role of $\rho_i$ is to capture linear correlations between the different fidelity levels. In general, $\rho_i$ can also be a function of $x$ leading to more complex schemes that can capture non-linear correlations \cite{perdikaris2017nonlinear}. Following \cite{kennedy2000predicting}, we complete our model specification by assuming independent Gaussian process priors for $\delta_i \sim \mathcal{GP}\,(\vec{0},k(\vec{x},\vec{x}'; \theta_i))$ and $f_1 \sim \mathcal{GP}\,(\vec{0},k(\vec{x},\vec{x}'; \theta_1))$. To simplify our presentation, and without loss of generality, we will assume $s=2$ levels of fidelity from now on: a low level, which we will denote with the subscript $L$ and a high level denoted with $H$. Now we have a composed dataset with the low and high fidelity levels: $\mathcal{D}= [\{(\vec{x}_{L_i}, y_{L_i})_{i=1}^{N_{L}}\}, \{(\vec{x}_{H_i}, y_{H_i})_{i=1}^{N_{H}}\}] = \{[\vec{X}_L, \vec{X}_H], [\vec{y}_L, \vec{y}_H]\} = \{\vec{X},\vec{y}\}$. As a consequence of the auto-regressive model we have chosen, we can write the joint prior distribution of the latent functions as \cite{kennedy2000predicting}:
\beq
\vec{f} =  \left[ 
  \begin{array}{c} \vec{f}_{L} \\ \vec{f}_{H} \end{array} \right] 
\sim \mathcal{N}\left(\left[\begin{array}{c} \vec{0} \\ \vec{0} \end{array} \right],  
\left[ \begin{array}{c c} \ten{K}_{LL} & \ten{K}_{LH}
 \\ \ten{K}_{LH}' & \ten{K}_{HH}
  \end{array} 
  \right]\right), 
\eeq
with
\beq
\begin{array}{l@{\hspace*{0.1cm}}c@{\hspace*{0.1cm}}
              l@{\hspace*{0.1cm}}l@{\hspace*{0.1cm}}
              c@{\hspace*{0.1cm}}l@{\hspace*{0.1cm}}l}
  \ten{K}_{LL} 
& = & & k_{L}&(\vec{X}_{L},\vec{X}_{L}';\theta_{L})  \\
  \ten{K}_{LH} 
& = &\rho & k_{L}&(\vec{X}_{L},\vec{X}_{H}';\theta_{L}) \\
  \ten{K}_{HH} 
& = &\rho^2 & k_{L}&(\vec{X}_{H},\vec{X}_{H}';\theta_{L}) 
& + & k_{H}(\vec{X}_{H},\vec{X}_{H}';\theta_{H})  \,.
\label{eq:MF_K}
\end{array}
\eeq
Now the entire covariance matrix for the different levels of fidelity $\ten{K}$ has a block structure, where $\ten{K}_{HH}$ and $\ten{K}_{LL}$ model the data in each fidelity level and $\ten{K}_{LH}$ models the cross-correlation between levels of fidelity, according to the priors specified in equations (\ref{eq:prior_f_i}) and (\ref{eq:prior_f_1}). We  also have kernel parameters for the different levels of fidelity. In principle, we could use different kernels for each level of fidelity, but we decide to use the squared exponential kernel, which results in parameters $(\eta_H, \ell_{H_1},\dots,\ell_{H_M})$, and $(\eta_L, \ell_{L_1},\dots,\ell_{L_M})$. For these parameters and the scaling factor $\rho$ we set the following priors:
\begin{eqnarray}
  \eta_H, \eta_L & \sim & {\rm HalfNormal}(\sigma = 5)\\
  \ell_{H_m}, \ell_{L_m} & \sim & {\rm Gamma}(\alpha = 2, \beta = 2), \ \ m = 1,\dots,M\\
  \rho & \sim & {\rm Normal}(\mu = 0, \sigma = 10)
\end{eqnarray}

We can perform inference and prediction for this model in the same way as for the single fidelity classifier, as detailed in Section~\ref{sec:sf}. In particular, we can use equations~(\ref{eq:posterior_mean}) and (\ref{eq:posterior_variance}) with the entire covariance matrix $\ten{K}$ to obtain the conditional mean and covariance of $\vec{f}^*$.

\subsection{Sparse Gaussian process classification}
\label{sec:sparse}

To increase the number of samples our can methodology handle efficiently, we introduce a sparse approximation of the Gaussian process prior. In particular, we replace the prior $f \sim \mathcal{GP}\,(\vec{0},k(\vec{x},\vec{x}'; \theta))$ with another function $u \sim \mathcal{GP}\,(\vec{0},k(\vec{x}_u,\vec{x}_u'; \theta))$ \cite{snelson2006sparse,quinonero2005unifying}. Here, we will have $M$ \textit{inducing points} $\vec{x}_{ui}$, which we group in a matrix $\vec{X}_u = \{(\vec{x}_{ui})_{i = 1}^M\}$. The number of inducing points $M$ needs to be smaller than the number of data points $N$ to gain in computational cost, while their role is to effectively summarize any redundancies in the observed data \cite{snelson2006sparse}. We can then compute our original latent variable $f$ with an independent covariance structure:
\begin{equation}
    f \sim \mathcal{N}(\ten{K}_{fu}\ten{K}_{uu}^{-1}\vec{u}, {\rm diag}(\ten{K}_{ff} - \ten{Q}_{ff}) + \sigma^2\ten{I}) \label{eq:sparse}
\end{equation}
Here, we introduce the following matrices that evaluate the kernel: $\ten{K}_{ff} = k(\vec{X}, \vec{X})$ and $\ten{K}_{uu} = k(\vec{X}_u, \vec{X}_u)$, and $\ten{Q}_{ff} = \ten{K}_{uf}^T\ten{K}_{uu}^{-1}\ten{K}_{uf}$, where $\ten{K}_{uf} = k(\vec{X}_u, \vec{X})$. The parameter $\sigma$ represents uncorrelated Gaussian noise in the regression setting. In this work, we deal with noise-free computer simulations and we set this parameter $\sigma = 0.1$ to help with the invertibility of the covariance matrix, which may be singular depending on the selected inputs \cite{neal1999regression}. The main computational advantage comes from the size reduction of the full covariance matrix $\ten{K}$, which is $N \times N$ to $\ten{K}_{uu}$, which is $M\times M$. The cost of inverting the full matrix is $\mathcal{O}(N^3)$ that we replacing by inverting a smaller matrix with cost $\mathcal{O}(M^3)$ and the trivial inversion of the diagonal covariance matrix in equation~(\ref{eq:sparse}). Now, the dominant operation in time complexity becomes the matrix multiplication needed to compute $\ten{K}_{fu}\ten{K}_{uu}^{-1}$, which is $\mathcal{O}(NM^2)$.

We can make predictions at new locations as described in Section~\ref{sec:sf} with a new conditional mean and covariance:
\begin{eqnarray}
    \mu_s(\vec{x}^*) &=& \ten{K}_{*u}\ten{\Phi} \ten{K}_{uf}\ten{\Lambda}^{-1}\vec{f}\\
    \Sigma_s(\vec{x}^*) &=& \ten{K}_{**} - \ten{Q}_{**} + \ten{K}_{*u}\ten{\Phi}\ten{K}_{*u}^T
\end{eqnarray}
with $\ten{K}_{*u} = k(\vec{x}^*, \vec{X}_u)$, $\ten{\Phi} = \left(\ten{K}_{uu} + \ten{K}_{uf}\ten{\Lambda}^{-1}\ten{K}_{uf}^T\right)^{-1}$, $\ten{\Lambda} = {\rm diag}(\ten{K}_{ff} - \ten{Q}_{ff}) + \sigma^2\ten{I} $, $\ten{K}_{**} = k(\vec{x}^*,\vec{x}^*)$, and $\ten{Q}_{**} = \ten{K}_{u*}^T\ten{K}_{uu}^{-1}\ten{K}_{u*}$. 

In the multi-fidelity case, we will have a composite vector of inducing points $\vec{X}_u = [\vec{X}_{Lu}, \vec{X}_{Hu}]$ and all the $\ten{K}$ matrix evaluations are assembled using the structure in equation~(\ref{eq:MF_K}). Since, in general, we will have few high-fidelity points, we decide to place the inducing points $\vec{X}_{Hu}$ at the same locations as the training data $\vec{X}_H$. For the low-fidelity inducing points, we use the k-means algorithm to select $M_L$ cluster centroids from the training inputs $\vec{X}_L$. Although we can, in principle, optimize the inducing point locations, most current approaches rely on variational inference \cite{titsias2009variational,hensman2015mcmc}, which we believe is beyond the scope of the present work.  

\subsection{Active learning}
Leveraging the posterior uncertainty estimates of Bayesian models, active learning strategies enable the judicious and cost-effective acquisition of new informative data with the goal of maximally enhancing the accuracy of the predictive model \cite{cohn1996active,mackay2003information}. Both the single- and multi-fidelity classifiers proposed in this work are well suited for active learning. Here, we attempt to estimate the next point to sample our computational model that will increase the accuracy of the classifier the most. In the machine learning community, there are multiple data acquisition policies postulated for this purpose. In the context of Gaussian process classification, we use a simple heuristic that balances exploration and exploitation trade-off, namely the trade-off of sampling at new locations to reduce uncertainty (exploration) versus sampling at regions that are most informative for refining the target classification boundary (exploitation) \cite{kapoor2007active}. This sequential refinement procedure starts by defining a set of candidate locations to sample $\ten{X}_{cand}$. A typical approach is to sample the points that are closest to the classification boundary, where the class probability is 0.5. Considering the sigmoid function that we use, this is equivalent to the point at which the latent function $f$ is closest to 0. However, this methodology is highly sensitive to the initial set of points used to train the classifier. For example, if there are multiple boundaries and the initial classifier only detected one, this heuristic is very unlikely to detect the additional classification regions. To solve this issue, we take advantage of the predictive posterior uncertainty of $f$. Specifically, in regions with a lot of training data, the posterior variance will be small, but as we move away from the training samples, the variance will increase. If we include this variance in our active learning scheme, we can also explore the parameter space, rather than exploiting the estimated boundary. Combining these two concepts, we solve the following minimization problem to select the next point to sample:
\begin{equation}
    \vec{x}^{new} = \argmin_{\vec{x} \in \vec{X}_{cand}} \frac{|\vec{\hat{\mu}(\vec{x})|}}{\sqrt{\hat{\Sigma}(\vec{x})}}
    \label{eq:AL}
\end{equation}
where $\hat{\mu}$ and $\hat{\Sigma}$ are the Monte Carlo estimates of the mean and variance of $f(\vec{x})$, since we are using the posterior distribution of the parameters to make predictions rather than point estimates. We solve the optimization problem in equation~(\ref{eq:AL}) by computing this metric for all candidates and selecting the one with the lowest score. Next, we evaluate the computer code at the parameters $\vec{x}^{new}$ to generate a new class label $y^{new}$ and add this input/output pair to the data-set. Finally, we re-train the classifier with the new information and repeat the active learning process until a certain number of iterations has been reached. In the multi-fidelity case, we only perform active learning in the highest level of fidelity and assume that the lower levels are inexpensive enough to be sampled as needed.

\section{Application to synthetic data}
\label{sec:synthetic}

\begin{figure}[ht]
    \centering
    \includegraphics[width = \textwidth]{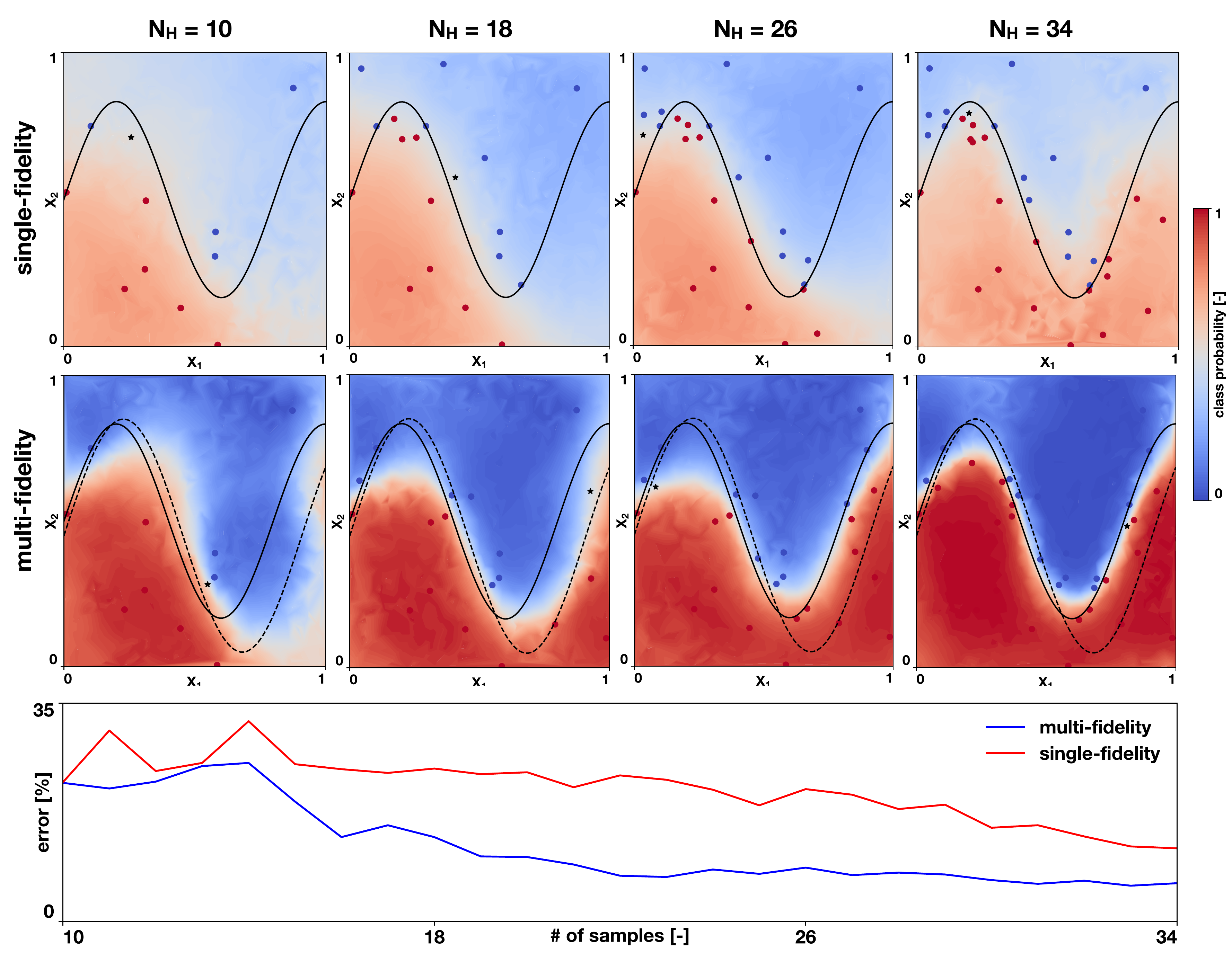}
    \caption{\textbf{Active learning of the synthetic example}. We define an arbitrary boundary to test the performance of the multi-fidelity classifier, as defined in equations~(\ref{eq:boundary_H}) and~(\ref{eq:boundary_L}). The high fidelity boundary is shown with a solid line and the low fidelity boundary is shown with a dashed line. For all steps the multi-fidelity classier presents a sharper boundary (middle row) and is more accurate (bottom row) than the single-fidelity classifier.}
    \label{fig:sine_example}
\end{figure}

We test the multi-fidelity classifier with a synthetic example in two dimensions. We define a low and high fidelity boundary with a sine function:

\begin{eqnarray}
    y_H(\vec{x}) &=& 
\begin{cases}
   1,  &\text{if }    0.5 + \sin(2.5\pi x_1)/3 - x_2 > 0 \\
    0,              & \text{otherwise}
\end{cases}\label{eq:boundary_H}\\
y_L(\vec{x}) &=& 
\begin{cases}
   1,  &\text{if }    0.45 + \sin(2.2\pi x_1)/2.5 - x_2 > 0 \\
    0,              & \text{otherwise}\label{eq:boundary_L}
\end{cases}
\end{eqnarray}
with $\vec{x} \in [[0,1],[0,1]]$. We note that the low fidelity boundary is a perturbed version of the high fidelity function, as can be seen in Figure~\ref{fig:sine_example}. We set the number of kernel length scales to 1 for this case for all classifiers. 
We generate a synthetic data set using the following procedure: First, we consider $N_L = 45$ low fidelity samples for the multi-fidelity classifier. To reduce the computational cost, we select 30 low fidelity samples that are close to the boundary and obtain the remaining 15 samples using a Latin hypercube sampling strategy \cite{stein1987large} to cover the remaining of the parameter space. For the sparse multi-fidelity classifier, we use the same $N_{L} = 45$ samples combined with $M_{L} = 30$ inducing points, which results in a reduction of computational cost. To select the high fidelity samples, we first obtain 500 low fidelity samples and randomly pick $N_H/2$ where $y_L = 1$ and $N_H/2$ where $y_L = 0$. This ensures that the initial samples are balanced between categories. We start the single-fidelity, multi-fidelity, and sparse multi-fidelity classifiers with the same samples to make fair comparisons. We test two cases: (i) We select $N_H = \{10,20,30,40,50\}$ samples as just described and run the classifier for each number 30 times for all three classifiers and (ii) We start with $N_H = 10$ and use active learning until we reach $N_H = 34$ and repeat this process 30 times for the single- and multi-fidelity classifiers. In both cases we test the accuracy of the classifiers using 1000 locations created with a Latin hypercube design. Finally, we compare whether one classifier is significantly more accurate than the other using non-parametric Wilcoxon signed-rank test \cite{wilcoxon1945individual}. To compare whether the active learning achieves better accuracy, we use the Mann Whitney U test \cite{mann1947test}. For both cases, we set the level of significance to 5\%.

\begin{figure}[t]
    \centering
    \includegraphics[width = \textwidth]{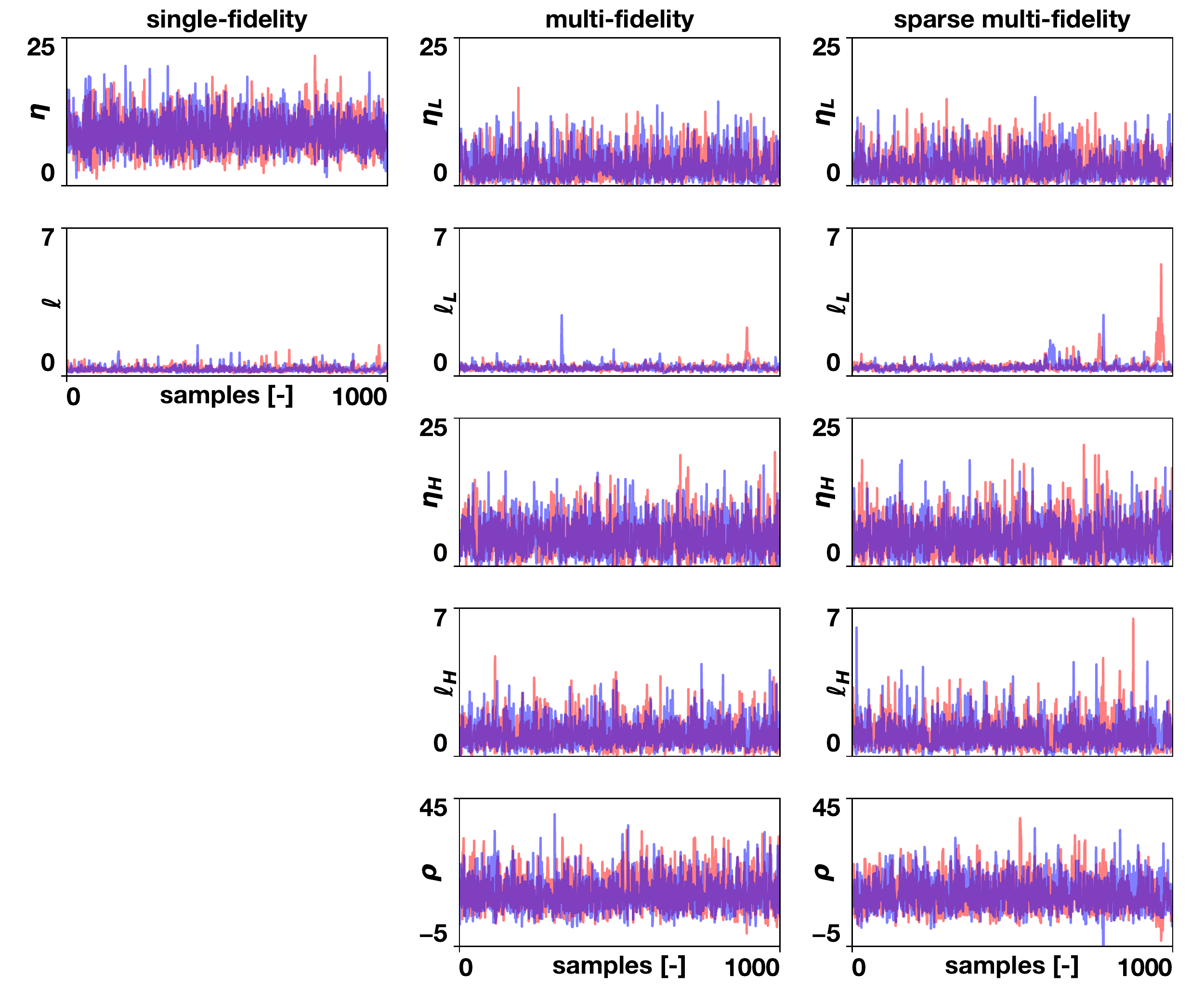}
    \caption{\textbf{Markov Chain Monte Carlo traces of the synthetic example.} We show the evolution of the parameter values as the sampler iterates for the single-fidelity, multi-fidelity, and sparse multi-fidelity classifiers. In all cases, the classifiers were trained with $N_H = 50$ and $N_L = 45$. The two chains are shown in red and blue for each parameter.}
    \label{fig:sine_trace}
\end{figure}

Figure~\ref{fig:sine_trace} shows trace plots depicting the evolution of the inferred parameters as the Markov Chain Monte Carlo sampler for all three classifiers trained with $N_H = 50$. The kernel parameters $\eta$ and $\ell$ exhibit similar behaviors and magnitudes for all classifiers. All parameters show similar patterns for the dense and sparse multi-fidelity classifiers. Remarkably, the high-fidelity length scale parameter $\ell_H$ is larger than its low-fidelity counterpart $\ell_L$, indicating that the details of the classification boundary are captured by the low fidelity Gaussian process $f_L$ and the differences of high and low fidelity are represented in the $\delta$ Gaussian process.

\begin{figure}[t]
    \centering
    \includegraphics[width = 0.8\textwidth]{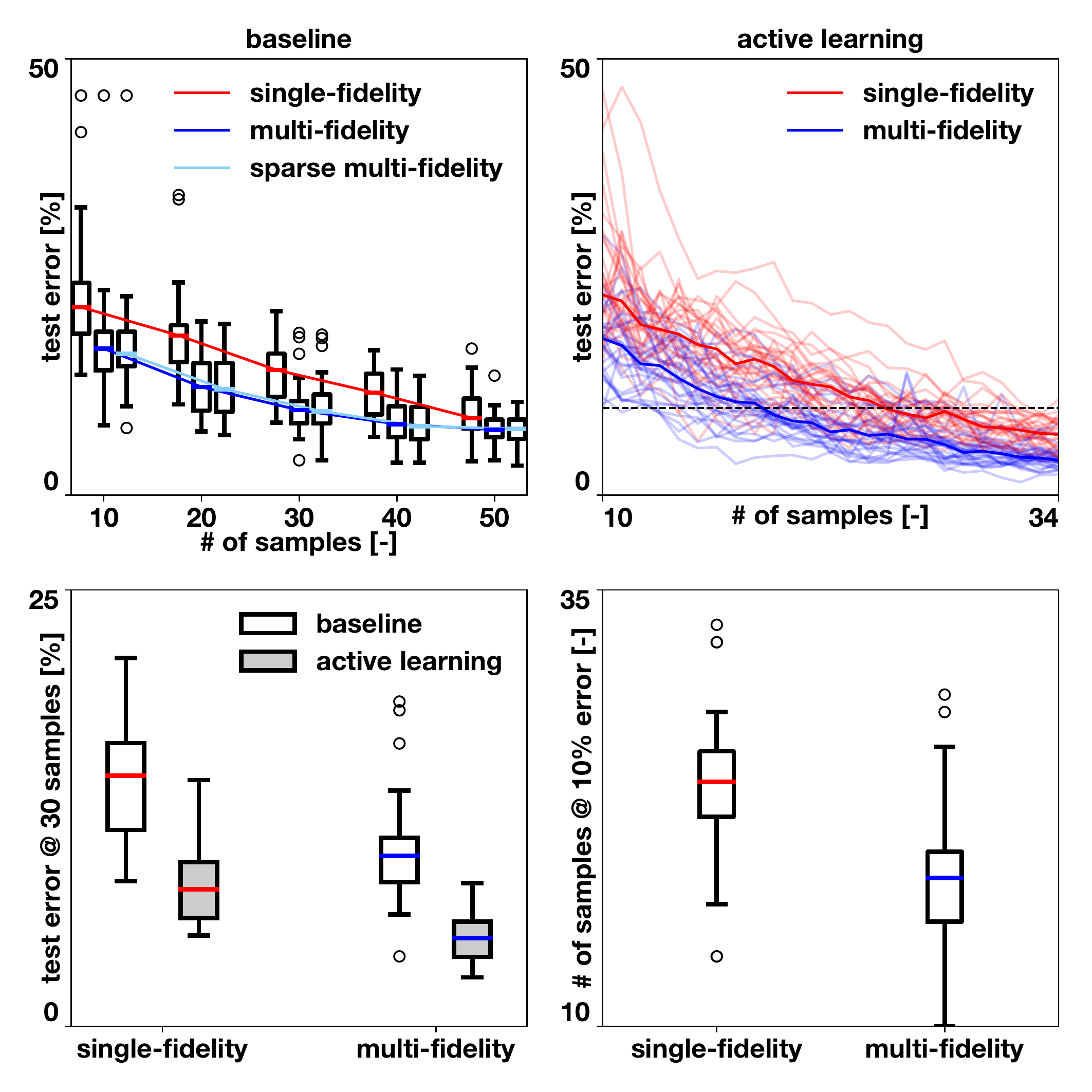}
    \caption{\textbf{Accuracy of the synthetic example.}  
The top left shows box plots for different numbers of high-fidelity samples with no active learning. The multi-fidelity classifiers always outperform the single-fidelity classifier. All differences are significant (p $<$ 0.01). The top right panel shows 30 active learning trajectories for each classifier type, where medians are displayed as solid lines. Both classifiers reduce their error when combined with the active learning strategy. We quantify this difference in the bottom left panel, where we compare the accuracy of the classifier trained with 30 samples with and without active learning. The active learning approach achieves significantly higher accuracy (p $<$ 0.001).  The bottom right panel quantifies the difference in accuracy between the single- and multi-fidelity classifiers by counting the number of samples required to achieve 10\% error when using active learning. The multi-fidelity classifier performs statistically better (p $<$ 0.005).}
    \label{fig:sine_accuracy}
\end{figure}

Figure~\ref{fig:sine_example} shows representative examples of the single- and multi-fidelity classifiers trained with active learning. At the beginning, the multi-fidelity classifier inherits the structure of the boundary that was learned from the low fidelity data. The single-fidelity classifier does not have this information and shows a poorer fit. The active learning strategy tends to reduce the error in both cases. However, this error reduction is achieved faster for the multi-fidelity classifier. To quantify the differences in accuracy, we first compare the classifiers without active learning (Figure~\ref{fig:sine_accuracy}, left). For sample sizes 10, 20, 30, 40 and 50, the multi-fidelity and sparse multi-fidelity classifiers are significantly more accurate than the single fidelity classifier (p $<$ 0.01). The multi-fidelity and sparse multi-fidelity classifiers behave similarly and there is no significant difference in accuracy for any sample size (p $>$ 0.16). The differences in median error range from 5.4\% at 10 samples to 1.3\% at 50 samples between the single fidelity and multi-fidelity classifiers. As the single-fidelity gains more information, the gap in accuracy is reduced. In the case of active learning, the trend is similar, where we see that the selected strategy reduces the error for both classifiers. Nonetheless, the multi-fidelity classifier has a higher median accuracy than the single-fidelity classifier. We quantify the effect of the the active learning strategy by comparing the accuracy of the case where we use the active learning strategy to achieve $N_H = 30$ to the case where we start the classifier with $N_H = 30$. The effect for both classifiers is a statistically-significant error reduction (p $<$ 0.001). The median error is decreased 7.1\% for the single-fidelity and 5\% for the multi-fidelity classifier. This difference can be explained because it is easier to reduce the error when it is higher. Finally, we quantify the differences in computational cost between the classifiers. For a target error of 10\%, the multi-fidelity classifier requires significantly less samples (p $<$ 0.005). The median number of samples is 24 for the single-fidelity classifier and 18.5 for the multi-fidelity classifier, representing a 23\% decrease in computational cost for this particular example.

\section{Application to cardiac electrophysiology}

\begin{figure}[t]
    \centering
    \includegraphics[width = \textwidth]{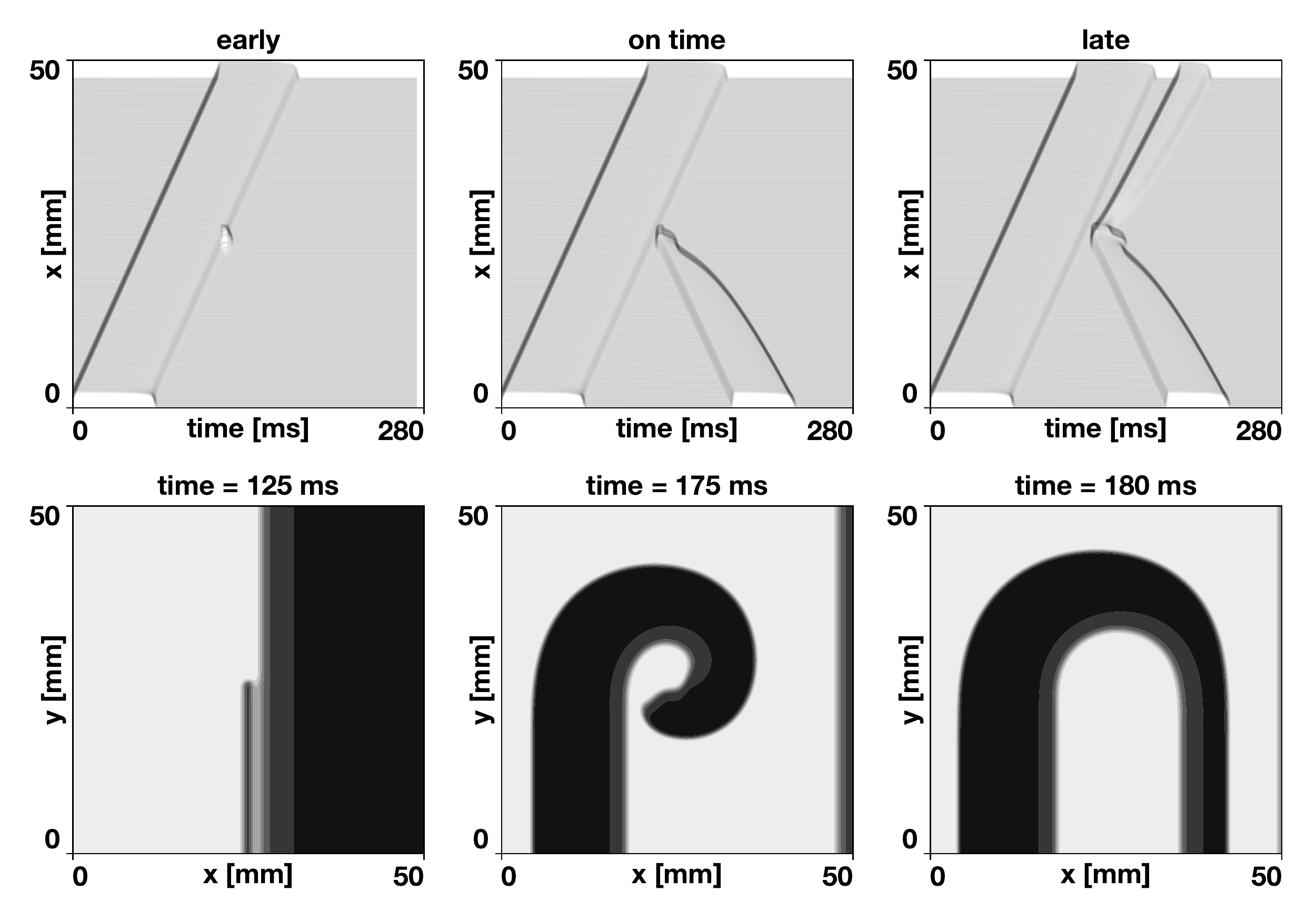}
    \caption{\textbf{Low and high fidelity models of the cardiac electrophysiology example.} The top row illustrates the low fidelity model, a one-dimensional cable, where the vertical axis represents the position on the cable, horizontal axis represents the time elapsed, and the lines represent the action potential. The bottom row shows the high fidelity model, a two-dimensional patch of tissue, where the contour plots represent the action potential at a given time, with black being activated and white represents the resting state. Panels on left shows a secondary stimulus that is applied too early to generate a spiral wave. Middle panels show a case when the stimulus is applied within the vulnerability window and a spiral wave is created. Right panels show an stimulus that is applied too late, failing to create a spiral wave.}
    \label{fig:cardiac_ep}
\end{figure}

\label{sec:cardic_ep}
In this section, we use the multi-fidelity classifier to study the formation of arrhythmic conduction in cardiac tissue. Cardiac tissue is susceptible to the formation of spiral waves \cite{gray1998spatial} when a secondary electrical stimulus is applied within certain time and location after a first planar wave passes. This range of time to successfully initiate a spiral wave is known as the vulnerability window \cite{Moreno2012}. It is hypothesized that, the larger this window, the greater the probability of arrhythmia for a given set of conditions including drugs or diseases. In the past, this quantity has been computed by brute-force in one-dimensional models of cardiac electrophysiology \cite{Moreno2012}. However, the window computed in one dimension might not be an accurate representation of what happens in two-dimensions or in the whole heart. As we will see in this example, it is possible that the spiral wave extinguishes as it meanders in the domain. Here, we will employ a multi-fidelity classifier to determine the vulnerability window in a two-parameter space, using the one-dimensional approximation as our low fidelity model. We choose a two-dimensional model instead of the full heart \cite{SahliCostabal2018a} to keep the computational cost tractable and maintain the ability to compute a test set for validating the accuracy of the classifier. 

To generate a spiral wave, after a primary wave is elicited, we apply a stimulus that results in a secondary wave that propagates in the opposite direction of the primary wave. This is illustrated in Figure~\ref{fig:cardiac_ep}, top row,  for the one-dimensional model. If we apply the stimulus too early (left), the primary wave will block it. If we apply the stimulus too late (right), a new wave will propagate in both directions. If we apply the stimulus within the vulnerability time-window, we obtain a single new wave that propagates in the opposite direction. In two dimensions, this results in a spiral wave, as can be seen in the bottom row of Figure~\ref{fig:cardiac_ep}.

Here, we use the monodomain equation with an Aliev-Panfilov cellular model \cite{Aliev1996} to characterize the electrical behavior of cardiac tissue. We use a domain size of 50 mm for the one-dimensional cable and 50x50 mm for the two-dimensional case. We discretize in time and space with finite differences, using a spatial step of 0.25 mm and a temporal step of 0.01 ms. We set the conductivity to 0.1 mm$^2$/ms. We solve the monodomain equation with an explicit time integrator implemented in FORTRAN \cite{Sahli2017}. To test spiral wave initiation, we first apply a stimulus on the left boundary to create a planar wave. Then, for 5 ms, we apply a second stimulus at a given time in the central 2 mm of the cable and in the central 2 mm of the bottom half of the two-dimensional tissue patch (Figure~\ref{sec:cardic_ep}, bottom left). Finally, in the one-dimensional model, we classify the spiral creation as successful if we detect an activation (dV/dt $>$ 0) at the left quarter but not at the right quarter of the cable after applying the second stimulus. We classify a successful spiral wave creation if we detect an activation (dV/dt $>$ 0) 300 ms after applying the second stimulus.

\subsection{Two-dimensional parameter space classification}
We begin by exploring two-dimensional parameter space, where we vary the time that we apply the second stimulus and the parameter $b$ of the Aliev-Panfilov model. We choose this particular parameter because we know that, besides controlling the action potential duration \cite{SahliCostabal2015,Hurtado2014}, it controls how much the spiral wave meanders in the domain. For some values of $b$, the wave will initially form, but it will instantly move to the boundaries of the domain, thus extinguishing itself. In these particular cases, we expect a disagreement in the labels between the low fidelity one-dimensional cable and the high fidelity two-dimensional simulation, causing the classification boundaries for the low and high fidelity data to be different. We set the range of the parameter $b$ to $\{0.035, 0.06\}$ and the time interval to $\{120,150\}$.

\begin{figure}[t!]
    \centering
    \includegraphics[width = \textwidth]{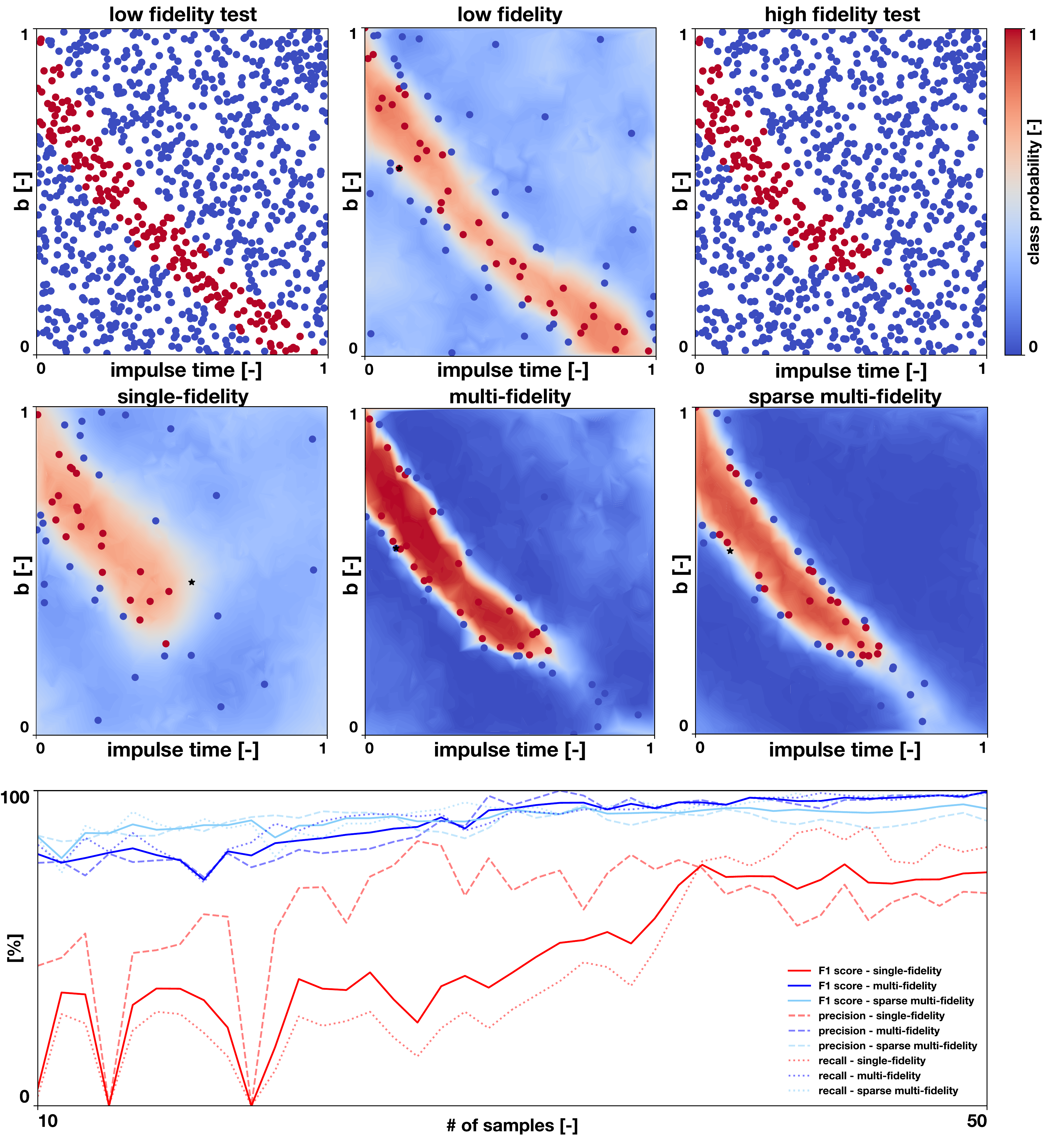}
    \caption{\textbf{Classifiers of the cardiac electrophysiology example.} Top left, test set used to evaluate the low fidelity classifier, with $N = 1000$. Top middle, single, low fidelity classifier trained with active learning to obtain $N_L = 84$ used in the multi-fidelity classifier. Top right, test set for the high-fidelity data. Middle left, resulting single-fidelity classfier with $N_H = 50$. Middle panel, resulting multi-fidelity classifier with $N_H = 50$. Middle right, sparse multi-fidelity classifier trained with $N_H = 50$. Bottom row, accuracy comparison between single-, multi-fidelity and sparse multi-fidelity classifiers, showing precision, recall, and F1 score. }
    \label{fig:cardiac_ep_results}
\end{figure}

In this example, we allow the kernel to have two different length scales $\ell$ with the same priors for both classifiers. To obtain the low fidelity samples for the multi-fidelity classifier, we train a single-fidelity classifier starting with 10 samples until we reach 84 samples with active learning. We then train a single- and a multi-fidelity classifier starting with $N_H = 10$ samples, where we select five samples from the low-fidelity samples with $y_L = 0$ and five samples with $y_L = 1$. We use active learning until we reach 50 samples and compute the precision, recall, and F1 score using a test set of $N_H = 1000$ created with a Latin hypercube design (Figure~\ref{fig:cardiac_ep_results}, top right). In this example, it is not appropriate to use accuracy to assess the performance of the classifiers, since only 13\% of the samples on the test belong to the class $y_H = 1$. If we used the accuracy as a metric, a classifier that only predicts $y_H = 0$ would already have an accuracy of 87\%. Precision is defined as the number of true positives divided by the sum of true and false positives. In other words, precision quantifies what proportion of the predicted labels $y_H = 1$ were correctly predicted as $y_H = 1$. Recall is defined as the number of true positives divided by the sum of true positives and false negatives. This metric quantifies proportion of the true labels $y_H = 1$ were correctly predicted as $y_H = 1$. Finally, the F1 score is the harmonic mean of precision and recall.

\begin{table}[h]
\centering
\begin{tabular}{|l|c|c|c|c|c|c|}
\hline
                & \multicolumn{2}{c|}{\textbf{precision {[}\%{]}}} & \multicolumn{2}{c|}{\textbf{recall {[}\%{]}}} & \multicolumn{2}{c|}{\textbf{F1 score {[}\%{]}}} \\
                \hline
\textbf{\# of samples  } & 10                 & 50                & 10               & 50               & 10                & 50                \\
\hline
single-fidelity & 44.4               & 67.5              & 3.0              & 82.1             & 5.6               & 74.1              \\
multi-fidelity  & 77.1               & 100               & 82.8             & 99.3             & 79.9              & 99.6    \\ 
sparse multi-fidelity  & 85.7               &    90.4            & 85.1             & 98.5             & 85.4              & 94.3    \\ 
\hline
\end{tabular}
\caption{\textbf{Accuracy of the cardiac electrophysiology example.} Precision, recall and F1 score are shown for single-, multi-fidelity and sparse multi-fidelity classifiers with the initial $N_H = 10$ samples and after 40 iterations of active learning, $N_H = 50$.} 
\label{tb:ep_accuracy}
\end{table}

\begin{figure}[ht]
    \centering
    \includegraphics[width = \textwidth]{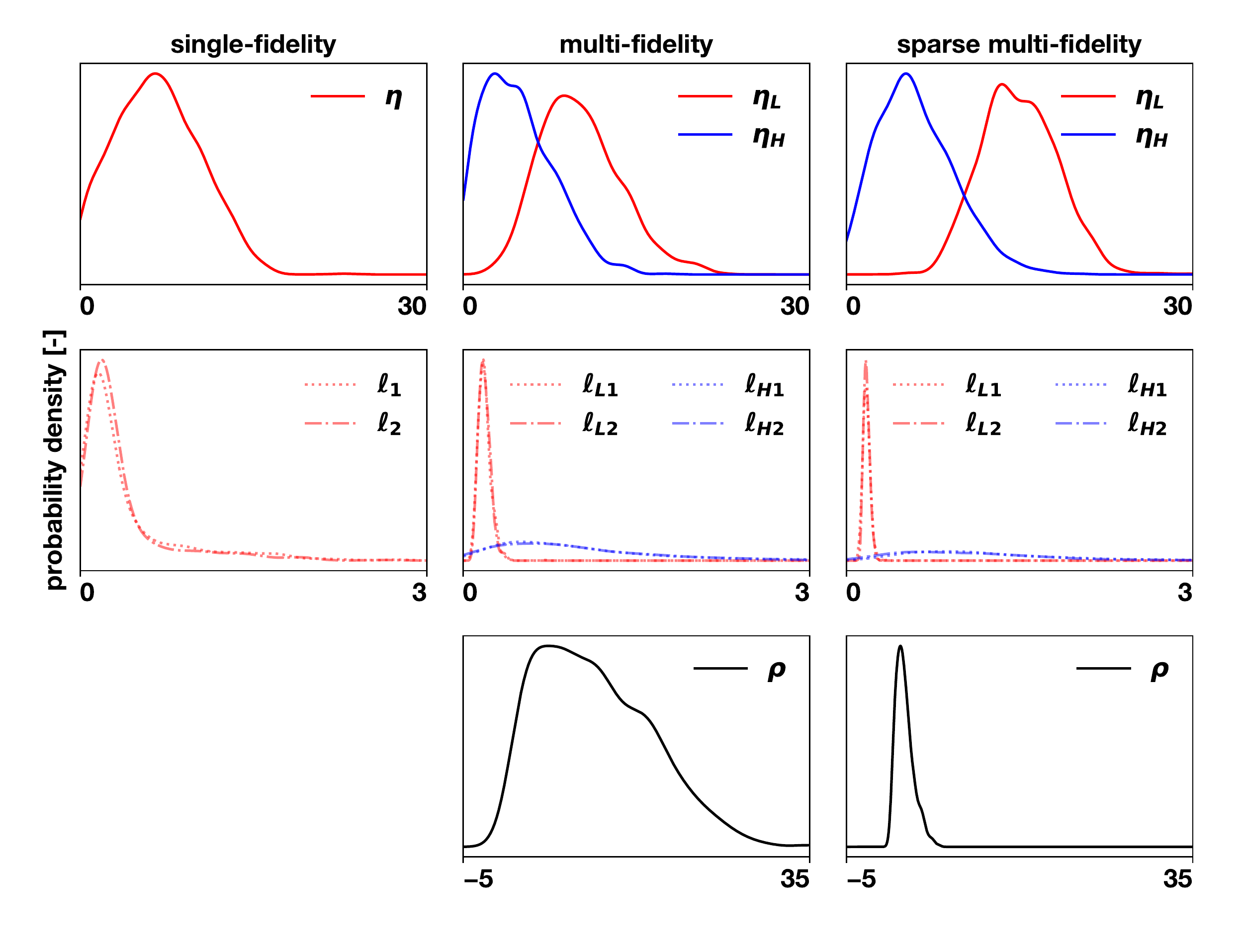}
    \caption{\textbf{Two-dimensional parameter space of cardiac electrophysiology application.} We show the distrubutions after the 3 classifiers have been trained with 40 rounds of active learning and $N_H = 50$.}
    \label{fig:posterior_ep}
\end{figure}

Our results, summarized in Figures~\ref{fig:cardiac_ep_results} and \ref{fig:posterior_ep} and Table~\ref{tb:ep_accuracy}, show that the multi-fidelity classifiers outperform the single-fidelity classifier. Qualitatively, we observe that both the multi-fidelity classifier and sparse multi-fidelity (Figure~\ref{fig:cardiac_ep_results}, middle and middle right) have placed most of the samples near the boundary, while the single-fidelity classifier (Figure~\ref{fig:cardiac_ep_results}, middle left) spend a significant portion of the samples exploring the entire parameter space. Remarkably, the active learning heuristic detected the difference in classification boundaries between the low fidelity (Figure~\ref{fig:cardiac_ep_results}, top middle) and the high fidelity data. Quantitatively, both multi-fidelity classifiers have better precision and recall with $N_H = 10$ than the single-fidelity classifier with $N_H = 50$, as seen in Table~\ref{tb:ep_accuracy} and Figure~\ref{fig:cardiac_ep_results}, bottom row. This trend continues as the number of samples acquired by active learning increase. The multi-fidelity classifier achieves 100\% precision and 99.6\% F1 score with $N_H = 50$. 

Figure~\ref{fig:posterior_ep} shows the posterior distribution for the parameters in all three classifiers after trained with 40 iterations of active learning. Similar to the synthetic data example, the low fidelity length scale parameters $\ell_L$ tend to be smaller than their high fidelity counterparts $\ell_H$, meaning that the detail of the classification boundary region is characterized by the low fidelity function. For the single-fidelity classifier, the length scale parameters $\ell$ tend to be larger than $\ell_L$, which results in a less defined boundary as can be seen in Figure~\ref{fig:cardiac_ep_results}, bottom left. Both multi-classifiers display similar distributions for all parameters. 

\subsection{Three-dimensional parameter space classification}
\begin{figure}[t]
    \centering
    \includegraphics[width = \textwidth]{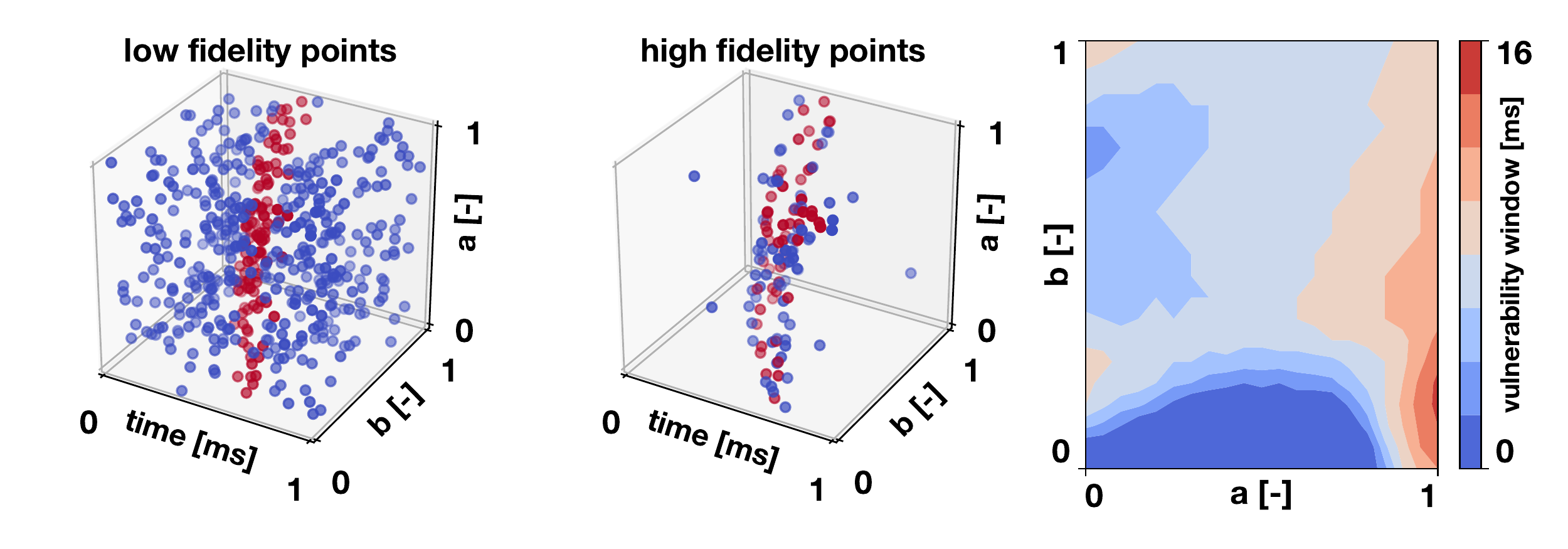}
    \caption{\textbf{Three-dimensional parameter space of cardiac electrophysiology application.} We additionally perturb the parameter $a$ of the Aliev-Panfilov model and detect the initiation of spiral waves in the two-dimensional tissue model. On the left, 509 low fidelity points used to train the sparse multi-fidelity classifier are shown. In the middle, we display 130 high-fidelity points after 100 iterations of active learning. On the right, we show the vulnerability window for combinations of $a$ and $b$, which represents the time available to induce a spiral waves after 100 iterations of active learning.}
    \label{fig:cardiac_ep_multidim}
\end{figure}

We study a three-dimensional parameter space by also varying the parameter $a$ of the Aliev-Panfilov, along with the parameter $b$ and the impulse time. This problem requires a large number of samples to characterize the low-fidelity boundary, therefore we use the sparse multi-fidelity classifier. We vary the parameter $a$ in the interval $\{0.1,0.2\}$, $b$ in $\{0.035,0.06\}$ and the time to apply the stimulus between 105 and 160 ms. We first compute 1000 low-fidelity samples with a Latin hypercube design to examine the boundary. Due to the small amount of $y_L = 1$ samples, we construct a dataset with all 109 $y_L = 1$ samples along with 400 $y_L = 0$ samples, resulting in $N_L = 509$. We then select $N_H = 30$, where 15 samples came from low-fidelity samples with $y_L = 0$ and 15 samples came from low-fidelity samples with $y_L = 1$. We use $M_L = 50$ inducing points for the low fidelity data. We train the classifier for 100 active learning iterations and compute the vulnerability window. This quantity is approximated by integrating the predicted class probability $y^*$ over the impulse time parameter. We approach this using a uniform grid spaced every 0.5 ms in 21x21 locations. The results are shown in Figure~\ref{fig:cardiac_ep_multidim}, where the low-fidelity points are displayed on the left, and the resulting high-fidelity points after 100 iterations of active learning are shown on the middle panel. Remarkably, most of the points of the high-fidelity model are located near the boundary, expending a small amount of the computational budget exploring the parameter space. Results on the vulnerability window are presented on the right of Figure~\ref{fig:cardiac_ep_multidim}, showing that for certain parameter combinations, this quantity is zero, where it is possible avoid the formation of spiral waves. Some other combination of parameters may provide a larger opportunity to induce a spiral wave.

\section{Discussion}
\label{sec:discussion}
In this work, we present a novel multi-fidelity classifier using Gaussian process priors. While the multi-fidelity paradigm has been proposed for regression \cite{kennedy2000predicting} and uncertainty quantification \cite{schiavazzi2017generalized}, this work represents one of the first attempts to formulate and implement a fully-Bayesian multi-fidelity classifier. Based on an autoregressive Gaussian process prior that enables the seamless integration of data with different levels of fidelity, our classifier outperforms the single-fidelity classifier both with synthetic data and in an application of cardiac electrophysiology. We adopt a fully Bayesian approach and set priors to the kernel hyper-parameters using efficient implementations of Markov-chain Monte-Carlo samplers \cite{salvatier2016probabilistic,hoffman2014no}. We also introduce a sparse approximation to extend the amount of data that our methodology can handle.

To strike a balance between classification accuracy, data efficiency, and computational cost we propose an active learning approach that takes advantage of the predictive uncertainty in our predictions \cite{kapoor2007active}. This approach significantly reduced the error in both examples shown here, showing a good combination of exploration of the parameter space and exploitation of the boundary.  Other authors have proposed sampling points of maximum information entropy \cite{Gramacy2017}; however, it is easy to see that the approach used here is similar. The entropy would be maximum for a uniform distribution, which, in our case, would be equivalent to a prediction of $f$ with zero mean and large variance. Such case would be close to minimize our active learning heuristic in equation~(\ref{eq:AL}). The actual minimum is for the case of zero mean and infinite variance, however the variance will be bounded for real applications. At the same time, the entropy will be minimum when all predictions are $y = 1$ or $y = 0$. In our implementation, this would correspond to a mean prediction of $f = \infty$ or $f = -\infty$, which maximizes our heuristic. 

 Although the multi-fidelity and sparse multi-fidelity classifier work better in all examples presented here, the differences are greater in the cardiac electrophysiology example. This suggests that multi-fidelity classifiers are advantageous when there is class imbalance: only a small region of the parameter space is labelled with a particular class. The multi-fidelity classifiers already have a candidate boundary that is encoded by the low fidelity data and it only needs to infer a simpler function $\delta$ to account for the differences in the low and high fidelity boundaries. In contrast, a single-fidelity classifier would spend most of its computational budget exploring the remaining parameter space. In the two-dimensional parameter space example, the low recall values for the single-fidelity classifier suggest that it is struggling to detect the $y_H = 1$ region, while the multi-fidelity classifiers have this information encoded in the low-fidelity data. The active learning training seems to be more effective in the multi-fidelity classifier than in the sparse version of it. Although the sparse multi-fidelity classifier starts with better accuracy at $N_H = 10$, it progresses at a slower pace than the full multi-fidelity classifier, achieving an F1 score 94.3\%. Since our active learning metric depends on the variance of the Gaussian process, this could be related to the independent covariance assumption used to reduce the computational cost in the sparse version. Overall, the cardiac electrophysiology application highlights the potential of this methodology. Combined with more complex electrophysiology models, this tool could assess the effect of drugs on the inducibility of arrhythmias, for example. 

Although we have successfully used our multi-fidelity classifiers in the examples presented here, our methodology suffers from some limitations that open new research directions. First, as we mention in Section~\ref{sec:sparse}, the full classifier scales with $\mathcal{O}(N^3)$ and the sparse version with $\mathcal{O}(NM^2)$, where $N$ is the number of data points and $M$ is the number of inducing points \cite{snelson2006sparse,titsias2009variational}. Additionally, using 
Markov-chain Monte-Carlo methods to perform the statistical inference is expensive. Even though the implementation we use here is highly efficient, training our classifiers took in the order of 10 minutes on a laptop for the two-dimensional example at $N = 95$, and up to two hours for the three-dimensional case that considered a sparse classifier with $N = 639$ and $M = 180$. This suggests that the methodology presented here will be useful when the time to compute the high fidelity model is on the order of several hours, which is not unusual. The sparse implementation allow us to drastically increase the number of data points  we can handle, but we are still bounded to a regime of at most $N = 1000$ to obtain reasonable training times, which is typical for Gaussian processes. A future direction to reduce the cost of training the classifier is to use variational inference \cite{Hensman2014,hensman2015mcmc} instead of Markov chain Monte Carlo. This methodology would allow us to significantly increase the number of samples we can handle at the cost of loosing expressiveness and interpretability of the Gaussian processes. Another area to improve our methodology is that currently, we need to retrain the classifier from scratch after each active learning iteration. An alternative could be to use the posterior distributions of the parameters as priors for the next iteration to accelerate the convergence of the Markov-chain Monte-Carlo sampler or to use a sequential Monte-Carlo sampler with online updates \cite{Gramacy2017}.

In summary, we propose a multi-fidelity classifier with active learning to efficiently predict the binary output of computer simulations. We envision that our new tool will help researchers to address problems that were prohibitively expensive without this approach.
%%%%%%%%%%%%%%%%%%%%%%%%%%%%%%%%%%%%%%%%%%%%%%%%%%%%%%%%%%%%%%%%%%%
\section*{Acknowledgments}
This work was supported by the School of Engineering Postdoctoral Fellowship from Pontificia Universidad Cat\'olica de Chile, the FONDECYT-Postdoctorado 3190355 awarded to F.S.C. and by the US Department of Energy under the Advanced Scientific Computing Research grant DE-SC0019116 to awarded P.P. This publication has received funding from Millenium Science Initiative of the Ministry of Economy, Development and Tourism of Chile, grant Nucleus for Cardiovascular Magnetic Resonance.
%%%%%%%%%%%%%%%%%%%%%%%%%%%%%%%%%%%%%%%%%%%%%%%%%%%%%%%%%%%%%%%%%%%

%%%%%%%%%%%%%%%%%%%%%%%%%%%%%%%%%%%%%%%%%%%%%%%%%%%%%%%%%%%%%%%%%%%

%%%%%%%%%%%%%%%%%%%%%%%%%%%%%%%%%%%%%%%%%%%%%%%%%%%%%%%%%%%%%%%%%%%
\bibliographystyle{elsarticle-harv}
\bibliography{litra}
%%%%%%%%%%%%%%%%%%%%%%%%%%%%%%%%%%%%%%%%%%%%%%%%%%%%%%%%%%%%%%%%%%%

\end{document}